\documentclass{article}

\usepackage{arxiv}

\usepackage[utf8]{inputenc} 
\usepackage[T1]{fontenc}    
\usepackage{hyperref}       
\usepackage{url}            
\usepackage{booktabs}       
\usepackage{amsfonts}       
\usepackage{nicefrac}       
\usepackage{microtype}      
\usepackage{lipsum}
\usepackage{graphicx}
\graphicspath{ {./images/} }

\usepackage{amsmath,mathtools}
\usepackage{float}
\usepackage{multirow}
\usepackage{siunitx}
\usepackage{subfig}

\title{Reinforcement Learning for\\Elliptical Cylinder Motion Control Tasks}

\author{
 Pawel Marczewski \\
  Institute of Computing Science\\
  Poznan University of Technology\\
  Poznan, 61-138 \\
  \texttt{pawel.marczewski@put.poznan.pl} \\
   \And
 Paulina Superczynska \\
  Institute of Automatic Control and Robotics\\
  Poznan University of Technology\\
  Poznan, 61-138 \\
  \texttt{paulina.superczynska@put.poznan.pl} \\
  \And
 Jakub Bernat \\
  Institute of Automatic Control and Robotics\\
  Poznan University of Technology\\
  Poznan, 61-138 \\
  \texttt{jakub.bernat@put.poznan.pl} \\
  \And
  Szymon Szczesny \\
  Institute of Computing Science\\
  Poznan University of Technology\\
  Poznan, 61-138 \\
  \texttt{szymon.szczesny@put.poznan.pl} \\  
}

\begin{document}
\maketitle
\begin{abstract}
The control of devices with limited input always bring attention to solve by research due to its difficulty and non-trival solution. For instance, the inverted pendulum is benchmarking problem in control theory and machine learning. In this work, we are focused on the elliptical cylinder and its motion under limited torque. The inspiration of the problem is from untethered magnetic devices, which due to distance have to operate with limited input torque. In this work, the main goal is to define the control problem of elliptic cylinder with limited input torque and solve it by Reinforcement Learning. As a classical baseline, we evaluate a two-stage controller composed of an energy-shaping swing-up law and a local Linear Quadratic Regulator (LQR) stabilizer around the target equilibrium. The swing-up controller increases the system's mechanical energy to drive the state toward a neighborhood of the desired equilibrium, a linearization of the nonlinear model yields an LQR that regulates the angle and angular-rate states to the target orientation with bounded input. This swing-up + LQR policy is a strong, interpretable reference for underactuated system and serves a point of comparison to the learned policy under identical limits and parameters. The solution shows that the learning is possible however, the different cases like stabilization in upward position or rotating of half turn are very difficult for increasing mass or ellipses with a strongly unequal perimeter ratio. 
\end{abstract}

\keywords{Reinforcement Learning \and Elliptical Cylinder \and Limited Torque \and Optimal Control \and Swing Up + LQR \and  Deep Q-Network (DQN)}

\section{Introduction}
The inspiration of this work is to show new control strategy of mechanical objects. Over the years, the mechanical properties has been exploited to obtain advanced systems like classical gearboxes \cite{shigleyMachines2010}, Origami art \cite{TaylorOrigami2011},  programming a mechanical motion by shape \cite{natureSobolev2023}. We are focused on the free objects that are driven by external forces or torques. For example, permanent magnets inside the human body are pulled by an external magnetic field to give motion of the laparoscopy camera \cite{Abbott:review:2020,Ebrahimi:2021}. This motion is continuous following the external force/torques. In this work, we propose a quasi-static solution to the control of mechanical objects. This means that we would like to follow from one state to another and stay stable. If we roll a ball with initial velocity, it will roll infinitely in the case of lack of friction. If we roll an elliptical cylinder it can turn over or oscillate around its position. In our work, we would like to use the shape of the object to obtain quasi-static position and get a controller from one to another. Furthermore, we consider an additional constraint, which is limited torque. This means that we cannot apply torque to freely rotate the object. This kind of problem is well known in the literature, for instance, the limited acceleration in an inverted pendulum problem. The reason for making such an assumption is to allow motion in even small external stimuli.

Reinforcement Learning (RL) has great potential in solving non-trival behavior and decision problems. It is particularly useful for solving tasks in which direct solutions are not easy to find. Over the years, the Reinforcement Learning has great popularity in the researchers community to solve difficult control problems: inverted pendulum, acrobot, mountain car continuous. These problems are popular benchmarks of RL in Gym environment \cite{pendulum_reward}. Current work demonstrates the application of the RL method to control the movement of Unmanned Aerial Vehicle (UAVs) \cite{UAV}, electric vehicles with reduced energy consumption \cite{ElectricVehicle}, underwater vehicles \cite{Underwater}, or connected vehicles with dynamic perception of traffic conditions \cite{ConnectedVehicles}. Contemporary RL systems can be very complex, e.g. they use the collective experience of a population of agents from many environments or parallel training using meta-networks \cite{RL_SOTA}.

In classical control, underactuated rotation tasks of rigid bodies are often addressed with hybrid controllers that combine a \emph{global} energy-based stage with a \emph{local} linear stabilizer. The energy-shaping (swing-up) component injects or dissipates mechanical energy to drive the system toward a neighborhood of the desired equilibrium, using a torque consistent with the sign of the energy error \cite{aastrom2000swinging,spong2002swing,ortega2002interconnection}. Once the state enters a region where linearization is valid, a local regulator such as Linear Quadratic Regulator (LQR) - obtained by solving the algebraic Riccati equation - provides fast damping and precise stabilization under bounded input \cite{anderson2007optimal,aastrom2021feedback}. In this work, we adopt this canonical energy-shaping + LQR scheme as a transparent and interpretable baseline for comparison with the reinforcement-learning policy under identical torque limits and model parameters.

In our work, we are inspired by an untethered magnetic device to define the control problem of an elliptical cylinder with limited input torque. Our task is to rotate and stabilize the object by applying the torque. This work defines the problem of control of an elliptical cylinder for four tasks. The main innovations are as follows: motion controller concept for elliptical cylinder with feedback by creating Reinforcement Learning problems related to mechanical model of elliptical cylinder.

The article is organised as follows. Section \ref{sec:related} presents related work in relation to modelling, learning methodology and control. Section \ref{sec:problem} presents the model of the controlled object and defines the problem of its control, dividing it into smaller tasks. The architecture of the trained model and the developed algorithms are presented in Section \ref{sec:model}. Section \ref{sec:experiments} presents the results of the experiments carried out for the defined tasks and compares them with the results obtained using the classical approach based on a two-stage controller. The work concludes with a summary in section \ref{sec:conclusions}.

\section{Related work}
\label{sec:related}

The related work is split into three main groups. The first according to modeling, the second related to Reinforcement Learning and the last on classical control approach.

In the field of modeling, the elliptical cylinder is considered as rigid body. This approach is well-known from mechanics \cite{landau1982mechanics}, however, the derivation of equation of energy and motion is not trival and is given in \cite{Linden2021}. In our work, we use these solutions to model motion of an object. The open loop example of elliptical cylinder in electromagnetic field with remote actuation can be seen in work \cite{chen2024rockingrollinghoppingexploring} or the problem of the magnetic torque production \cite{Abbott:review:2020,Bernat:KKR:2025,Ebrahimi:2021}. In future, these concepts can be inspiration to built experimental laboratory setup.

Reinforcement Learning has been proven to achieve positive results in the magnetic control of advanced physical systems in \cite{Degrave2022}. Considering that our problem has just one degree of freedom in 2-dimensional space, we decided to use the basic Deep Q-Network (DQN) method inspired by \cite{Mnih2015}. Despite the discrete output, DQN was successfully used for fine control in the Inverse Pendulum \cite{Israilov2023} and Cart-Pole Problems \cite{8125811}

Underactuated rotation and inversion problems are commonly solved with two-stage designs: an energy-based swing-up controller that drives the system toward the target energy level, followed by local linear stabilization around the desired equilibrium. This paradigm is well established for pendulum-like systems, including rotary and cart-pole configurations \cite{aastrom2000swinging,spong2002swing}. In the second stage, linearization around the goal enables the use of LQR (via the Riccati equation) to achieve fast convergence and robustness in the vicinity of the equilibrium \cite{anderson2007optimal,aastrom2021feedback}. Passivity-based formulations provide a systematic interpretation of energy shaping through modification of the closed-loop energy landscape \cite{ortega2002interconnection}. In our study, we use the swing-up + LQR controller as an interpretable reference method for the elliptical-cylinder task under the same torque constraints as in the RL setting.

\begin{figure}
\centering
\includegraphics[width=0.8\textwidth]{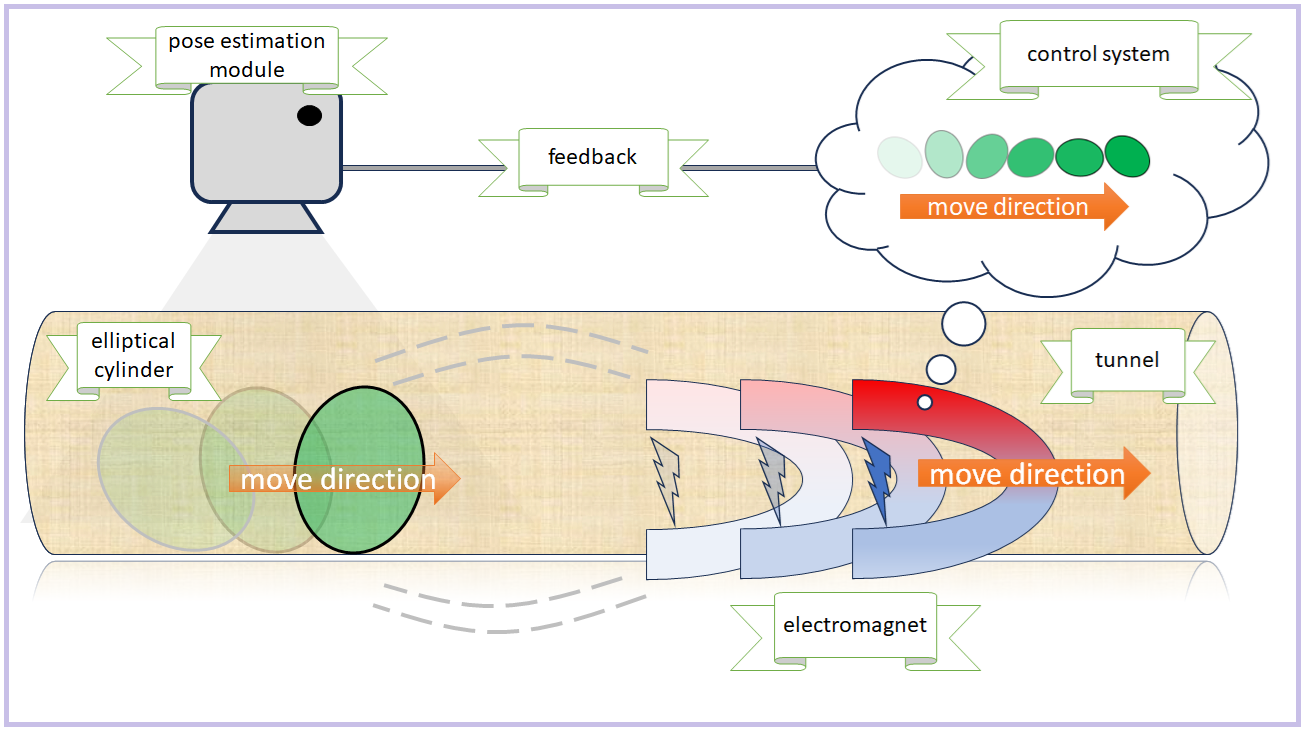}
\caption{Concept of an elliptical cylinder control system.}
\label{fig:system}
\end{figure}

\section{Problem formulation}
\label{sec:problem}
The general concept of the control system is shown in Fig. \ref{fig:system}. A magnetic elliptical cylinder is placed inside a tunnel made of any material that does not shield the magnetic field e.g. rubber hoses, plastic pipes or vessels. The cylinder is set in motion by a moving electromagnet with controlled magnetic force torque. The trajectory of movement is predefined by the system user. The system decides on the movement of the electromagnet and the direction of magnetic interaction based on the defined trajectory and feedback data that estimates the position of the cylinder. Solving this control problem requires first identifying the cylinder model and decomposing the continuous motion problem into position change tasks which are described in following subsections.

\subsection{Elliptical Cylinder Model}

As mentioned, this work considers the control problem of an elliptical cylinder by torque. The goal of the controller is to roll the cylinder on the plane without incline. The cylinder model is found by analyzing the kinetic and potential energy as in work \cite{Linden2021} and adding the control torque as input. We consider the following equations:
\begin{equation}
   J(\theta) \ddot{\theta} = \tau - \tau_{p}(\theta) - \tau_{f}(\theta, \dot{\theta})
\end{equation}
where $\theta$ is the angular position of the cylinder, $\tau$ is the input torque, $\tau_{p}$ is the torque related to the potential energy, and $\tau_{f}$ is the fictitious torque. The model parameters are given by: $m$, which is the mass, $I = m\frac{a^2 + b^2}{4}$, which is inertia moment, $a$ and $b$ which are the semi-major and semi-minor axes ($a \geq b$) respectively. The angular position $\theta$ is $0$ when semi-minor axes $b$ is parallel to rolling surface (horizontal alignment), and $\theta$ is $\pi/2$ when $a$ semi-major axis is perpendicular to rolling surface (vertical alignment). The definition of torques $\tau_p(\theta)$, $\tau_f(\theta,\dot{\theta})$ are equal to:
\begin{equation}
    \begin{aligned}
   \tau_{p}(\theta) &= mg \frac{\left(\frac{a}{b} - \frac{b}{a}\right)\alpha(\theta)\beta(\theta)}{\sqrt{\alpha^2(\theta) + \beta^2(\theta)}}  \\
   \tau_{f}(\theta, \dot{\theta}) &= m \frac{ab\left(a^2 - b^2\right)\alpha(\theta)\beta(\theta)}{\left[ \alpha^2(\theta) + \beta^2(\theta) \right]^2} \dot{\theta}^2
    \end{aligned}
\end{equation}
and $J(\theta)$ is:
\begin{equation}
J(\theta) = I + m \frac{a^2 \alpha^2(\theta) + b^2 \beta^2(\theta)}{\alpha^2(\theta) + \beta^2(\theta)}
\end{equation}
where: $\alpha(\theta) = a \text{sin}(\theta)$ and $\beta(\theta) = b \text{cos}(\theta)$ and $g$ is the gravitational acceleration.

\subsection{Problem definition}

\begin{figure}[htbp]
     \centering
        \includegraphics[width=0.5\textwidth]{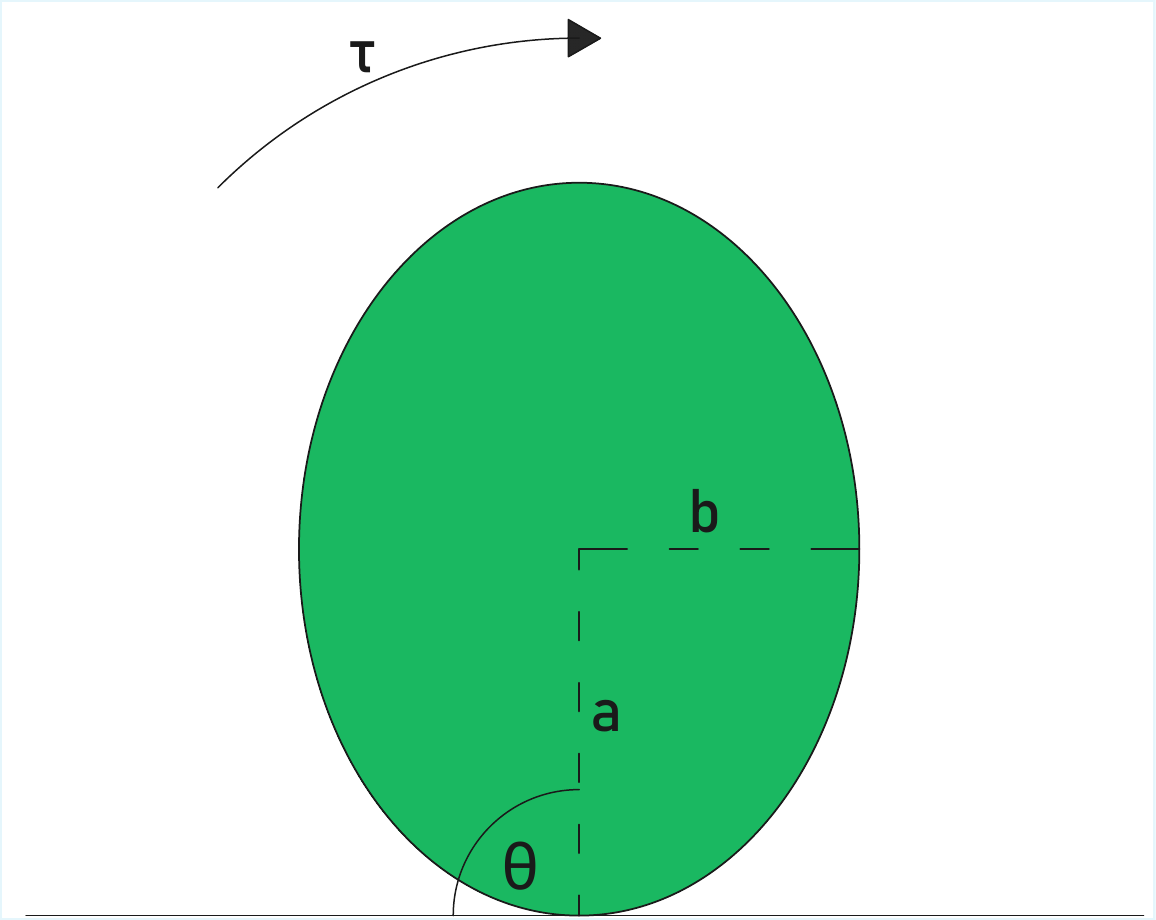}
    \caption{Cross-section of an elliptical cylinder for vertical alignment with $\theta = \frac{\pi}{2}$. }
    \label{fig:ellipse}
\end{figure}

The subject of the study is an elliptical cylinder, whose elliptical cross-section is shown in Fig. \ref{fig:ellipse}. The figure shows the semi-axes $a$ and $b$ of the ellipse, the current angle of rotation $\theta$, the angular velocity $\dot\theta$, and the current direction of torque $\tau$. 

\begin{figure*}[htbp]
     \centering
     \subfloat[][Task: $\frac{\pi}{2} \to \frac{3\pi}{2}$]{
        \includegraphics[width=0.55\textwidth]{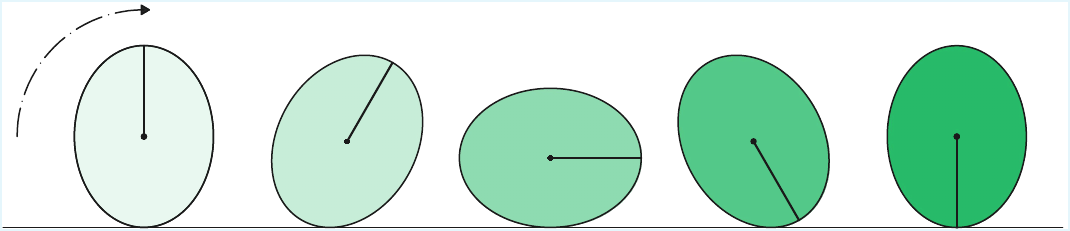}
        } 
     \subfloat[][Task: $0 \to \frac{\pi}{2}$]{
        \includegraphics[width=0.35\textwidth]{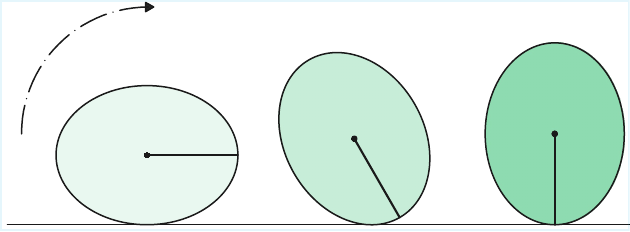}
        } \\
     \subfloat[][Task: $0 \to \pi$]{
        \includegraphics[width=0.55\textwidth]{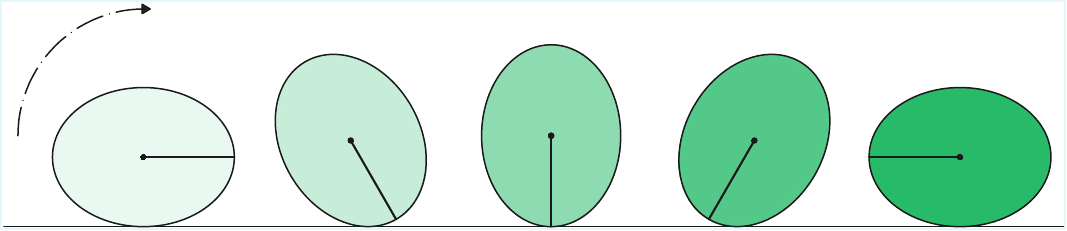}
        }
     \subfloat[][Task: $\frac{\pi}{2} \to \pi$]{
        \includegraphics[width=0.35\textwidth]{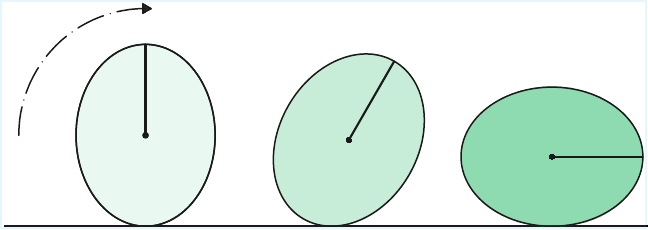}
        }
     \caption{Four scenarios representing four movement control tasks.}
     \label{fig:scenarios}
\end{figure*}

For the problem of controlling an elliptical cylinder, four tasks were defined to reflect possible movement scenarios:
\begin{enumerate}
  \item task: $\frac{\pi}{2} \to \frac{3\pi}{2}$, which means vertical to vertical rotation by an angle $\Delta\theta=\pi~rad$, as shown in Fig. \ref{fig:scenarios}a) 
  \item task: $0 \to \frac{\pi}{2}$, which means horizontal to vertical rotation by an angle $\Delta\theta=\frac{\pi}{2}~rad$, as shown in Fig. \ref{fig:scenarios}b) 
  \item task: $0 \to \pi$, which means horizontal to horizontal rotation by an angle $\Delta\theta=\pi~rad$, as shown in Fig. \ref{fig:scenarios}c) 
  \item task: $\frac{\pi}{2} \to \pi$, which means vertical to horizontal rotation by an angle $\Delta\theta=\frac{\pi}{2}~rad$, as shown in Fig. \ref{fig:scenarios}d) 
\end{enumerate}

\section{Model and training algorithm}
\label{sec:model}
\begin{figure}
\begin{center}
\includegraphics[width=0.6\textwidth]{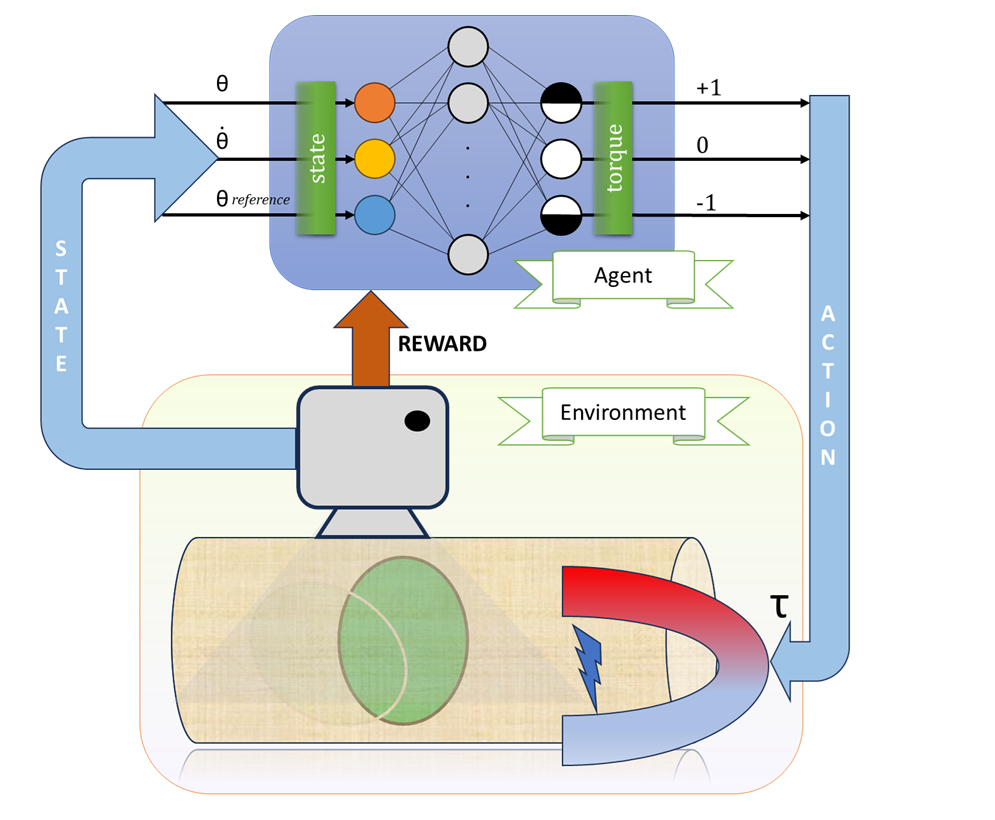}
\caption{Data flow diagram between the agent and the environment in the control system.}
\label{fig:environ}
\end{center}
\end{figure}

Fig.~\ref{fig:environ} describes the data flow between the agent and the controlled system. The agent receives information about the status: angle $\theta$, speed $\dot\theta$, and target angle $\theta_{ref}$; it then provides an output in the form of torque, which is applied to the environment by means of an electromagnet. The agent's output reflects three possible situations: force acting in one of two opposite directions  ($\tau_{}=10^{-3}\,\mathrm{Nm}$ or $\tau_{}=-10^{-3}\,\mathrm{Nm}$) or no force ($\tau_{}=0\,\mathrm{Nm}$). The reward function is determined based on the tracked position, speed, and torque applied to the elliptical cylinder. For every task, a separate model is trained. The following subsections present the model architecture, training strategy, and reward function selection policy.
\subsection{Reinforcement learning model}

The proposed agent is based on the classical \textbf{Deep Q-Network (DQN)} architecture, enhanced with mechanisms that improve training stability and learning efficiency. The implementation was developed in \texttt{Python}, using \texttt{PyTorch} for neural network modeling and \texttt{Gymnasium} as the environment interface.

\subsection{Agent Architecture}

The agent employs two identical neural networks:
\begin{itemize}
    \item an \textbf{online network}, responsible for estimating the action-value function $Q(s, a)$;
    \item a \textbf{target network}, used to compute stable target values during parameter updates.
\end{itemize}

Each network takes as input a 3-dimensional state vector ($\theta,\dot\theta, \theta_{\text{ref}}$ ), where $\theta_{\text{ref}}$ is the reference (target) angle. Neural Network (NN) is built with three fully connected hidden layers of size 64, each followed by a \textbf{ReLU} activation function. The final fully connected output layer produces the $Q$-values for all 3 possible actions. During inference, an action is taken based on the \textbf{argmax} function. Before training, the NN is initialized with pseudo-randomly generated weights based on a given seed for experimental reproducibility. 

\subsection{Exploration Strategy}

During the learning process, an \textbf{epsilon-greedy} exploration policy was used, where the agent selects a random action with probability $\epsilon$ and the action maximizing $Q(s, a)$ with probability $1 - \epsilon$. The exploration rate decays exponentially according to:
\begin{equation}
\epsilon_{t+1} = \max(\epsilon_{\text{min}}, \epsilon_t \cdot \epsilon_{\text{decay}})
\end{equation}
with $\epsilon_0 = 1.0$, $\epsilon_{\text{min}} = 0.05$, and $\epsilon_{\text{decay}} = 0.995$.

\subsection{Reward function}
The reward function is inspired by the pendulum environment taken from the Gymnasium library \cite{pendulum_reward}; it was designed to penalize deviations from the reference angle, excessive angular velocity, and large control torques. 

\begin{equation} \label{eq:reward1}
r = 
\begin{cases}
\mathmakebox[2em][l]{2,} \text{if } |\theta - \theta_{\text{ref}}| < 0.01 \text{ and } |\omega| < 1 \\
\mathmakebox[2em][l]{1,} \text{if } |\theta - \theta_{\text{ref}}| < 0.05 \\
\mathmakebox[15em][l]{- \Big[ (\theta - \theta_{\text{ref}})^2 + 0.1 \, \omega^2 + 0.001 \, \tau^2 \Big],} \text{otherwise}
\end{cases}
\end{equation}

\begin{figure*}[htbp]
     \centering
     \subfloat[][Reward function without distance normalization]{
        \includegraphics[width=0.5\textwidth]{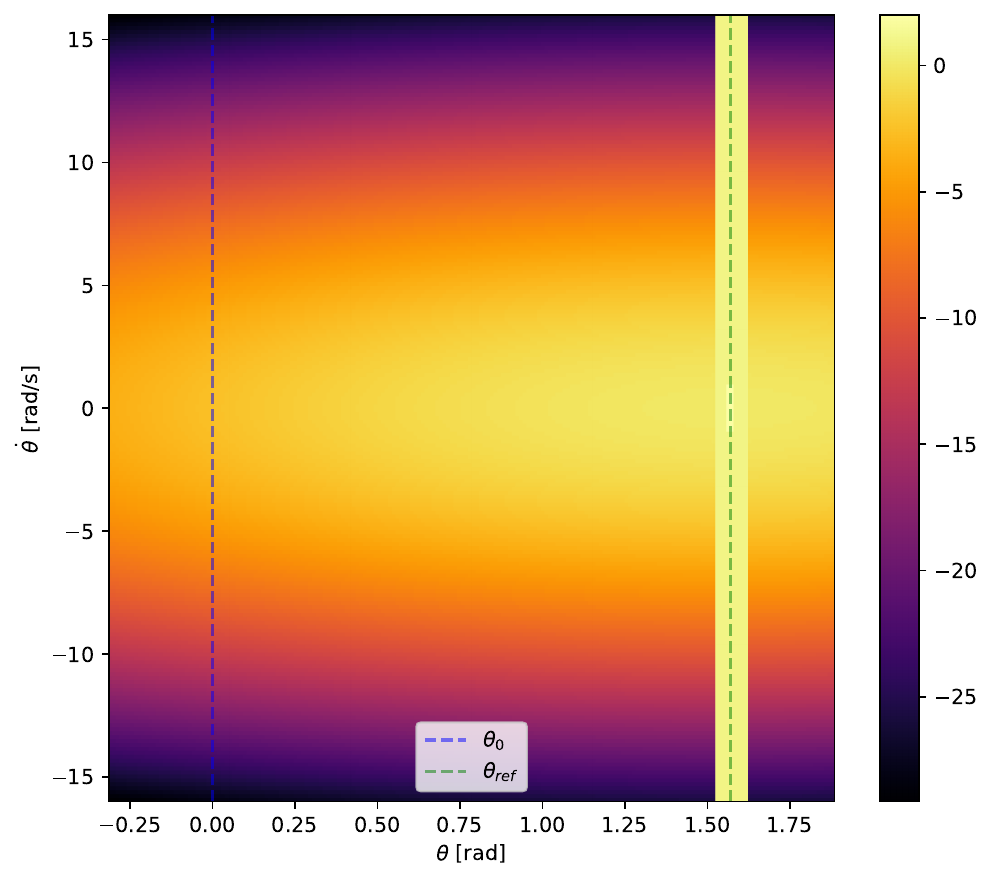}
        } 
     \subfloat[][Reward function that normalizes distance]{
        \includegraphics[width=0.5\textwidth]{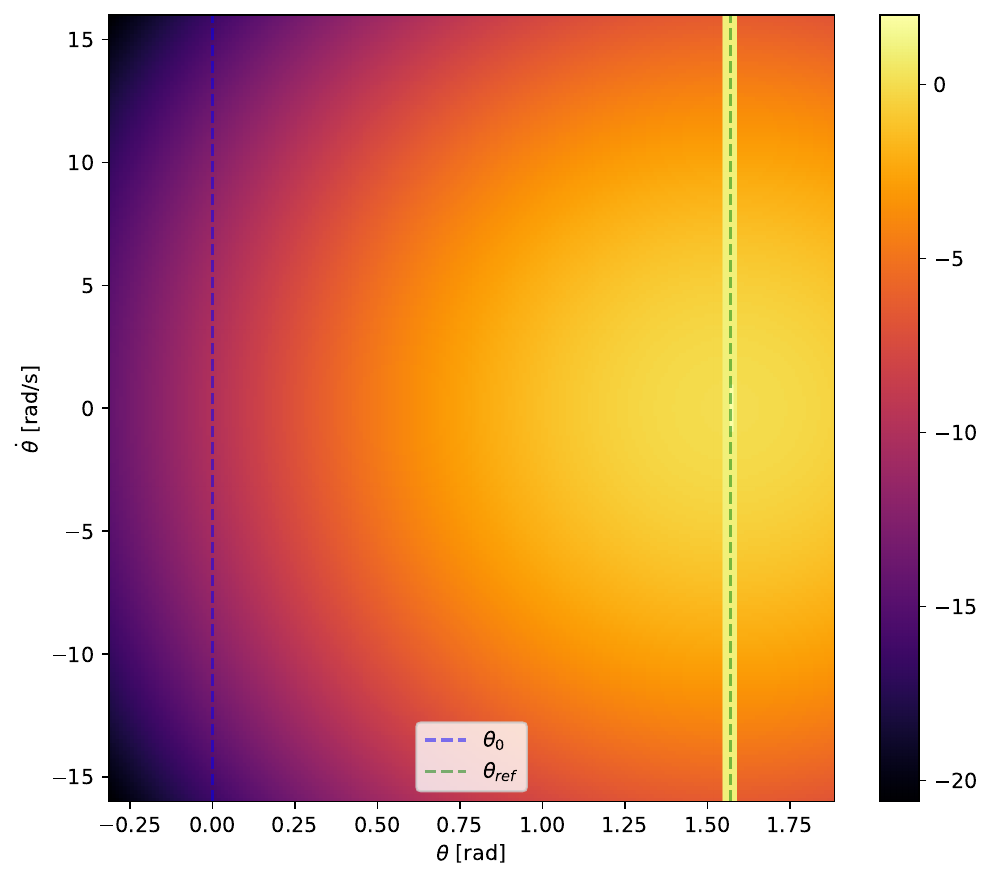}
        }
     \caption{Heatmaps of two reward functions for task $0 \to \frac{\pi}{2}$.}
     \label{fig:reward_heatmaps}
\end{figure*}

The adjusted reward function was used, which takes into account the amount of rotation that the agent needs to make by scaling the parameters in the reward function according to the size of the observation space, so that the reward function has a similar value distribution independent of the scenario. It is defined as:

\begin{equation} \label{eq:reward2}
r = 
\begin{cases}
\mathmakebox[2em][l]{2,} \text{if } |\theta - \theta_{\text{ref}}| < 0.001 \text{ and } |\dot{\theta}| < 0.001 \\
\mathmakebox[2em][l]{1,} \text{if } |\theta - \theta_{\text{ref}}| < 0.02 \\
\mathmakebox[21.3em][l]{- \left[
\left( \frac{\pi}{\theta_{\text{ref}} - \theta_0} \, (\theta - \theta_{\text{ref}}) \right)^2
+ 0.025 \cdot  \omega^2
+ 4000 \cdot \tau^2
\right],} \text{otherwise}
\end{cases}
\end{equation}
where $\theta$ denotes the current angular position, $\theta_{\text{ref}}$ the reference (target) angle, $\theta_0$ the initial angle  in radians, $\omega$ the angular velocity in $\frac{rad}{s}$, and $\tau$ the applied control torque in $Nm$. The difference between the reward functions (\ref{eq:reward1}) and (\ref{eq:reward2}) clearly presents Fig. \ref{fig:reward_heatmaps}.

\subsection{Experience Replay and Network Updates}
\label{replay}
A \textbf{replay buffer} was employed to store past transitions and enable random mini-batch sampling during learning, which reduces sample correlation and improves convergence stability. The network parameters were updated periodically every 10th step using \textbf{backward propagation} , utilizing the \textbf{temporal-difference (TD) error}, with the loss function defined as:
\begin{equation}
\mathcal{L} = \left( r + \gamma \cdot \max_{a'} Q_{\text{target}}(s', a') - Q(s, a) \right)^2
\end{equation}
where $r$ is the immediate reward, $\gamma = 0.99$ is the discount factor, $Q_{\text{target}}$ is the value predicted by the target network, and $Q$ is the value estimated by the online network. The optimization is performed using the \textbf{Adam optimizer} with a learning rate of $\alpha = 0.001$.

\section{Experiments}
\label{sec:experiments}
In tasks $\frac{\pi}{2} \to \frac{3\pi}{2}$  and $\frac{\pi}{2} \to \pi$, an elliptical cylinder starts with the advantage of greater potential energy because the center of mass is elevated. The task $\frac{\pi}{2} \to \frac{3\pi}{2}$ could theoretically be accomplished by applying a small amount of force at the beginning and an equal force in the opposite direction near the target, since the system can largely exploit its existing potential energy. In contrast, the task $\frac{\pi}{2} \to \pi$ requires the controller to counteract the potential energy difference, as the system must dissipate the stored energy to stop at the horizontal position. For those reasons, we focused on tasks $0 \to \frac{\pi}{2}$ and $0 \to \pi$ during experiments. They require more refined control in order to gain potential energy and move from a horizontal stable position and therefore require more precise control to inject energy into the system. In task $0 \to \frac{\pi}{2}$, the agent needs to stabilize the elliptical cylinder at the unstable vertical position. In the task $0 \to \pi$, the agent needs to cancel out the stored energy that was required to move to a vertical position in order to prevent overshooting and stabilize at the targeted horizontal position.

During inference, the simulation of an object was sampled at 1000 Hz, and the controller response was at 100 Hz in order to partially simulate inertia. Training for 1000 epochs with 2048 steps each took, on average, $1.49 \pm 0.15\,\mathrm{h}$
 on a Dell Latitude 5420 laptop with an Intel Core i5-1145G7 CPU. 

If not stated otherwise, simulations took place with the default values of the environment where $m=20\ g$, $a=34\ mm$, and $b=26\ mm$, and the model was trained with a batch size of 1024. 

\subsection{Basic Example}
\begin{figure}[htbp]
\begin{center}
\includegraphics[width=0.95\textwidth]{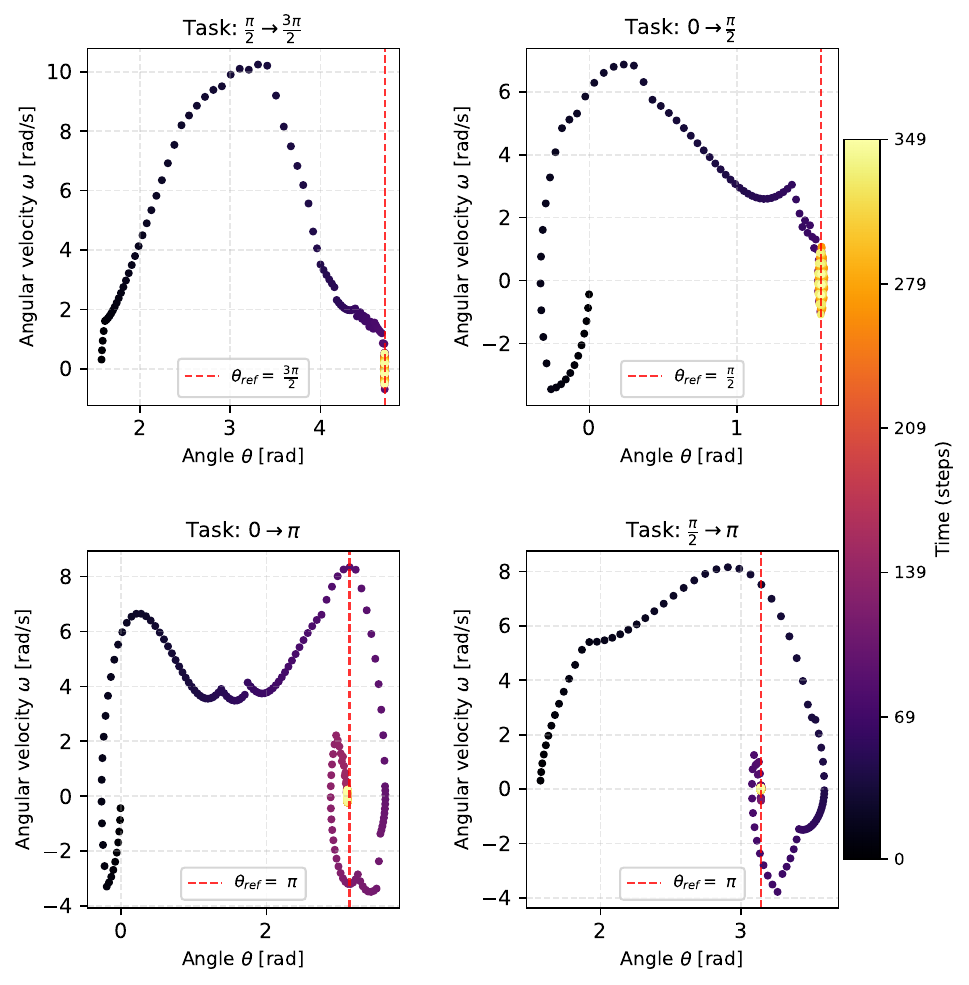}
\caption{Phase–space trajectories ($\theta$ vs.\ $\omega$) for all four tasks using trained agents. Each subplot illustrates the evolution of the system state, where $\theta$ denotes the angular position of the elliptical cylinder and $\omega$ its angular velocity. The color of the points represents consecutive time steps within a single simulation. The red dashed line indicates the reference angle $\theta_{ref}$.}
\label{fig:phase_space_trajectories_rl}
  
\end{center}
\end{figure}

As can be seen in Fig. \ref{fig:phase_space_trajectories_rl}, all four tasks were successfully completed by trained models; for each of them, a slightly different steering policy was adopted. In task $\frac{\pi}{2} \to \frac{3\pi}{2}$, the cylinder simply speeds up in the first half of the path and then slows down to accomplish the target. For tasks $0 \to \frac{\pi}{2}$ and $0 \to \pi$, the agent needs to take a different path and steer the cylinder in the opposite direction from the target to accumulate potential energy and, therefore, gain speed to travel from a stable horizontal position to a vertical one, in which the agent for task $0 \to \frac{\pi}{2}$ remains by slowing down and stabilizing. However, for the $0 \to \pi$ task, it goes slightly past the target before returning and staying at the target angle. For the  $\frac{\pi}{2} \to \pi$ task, a similar policy was applied; it just needed to gain speed from angle $\frac{\pi}{2}$ in comparison to task $0 \to \pi$.

\subsection{Influence of Cylinder Parameters}

\begin{figure}[htbp]
\centering
\includegraphics[width=0.95\textwidth]{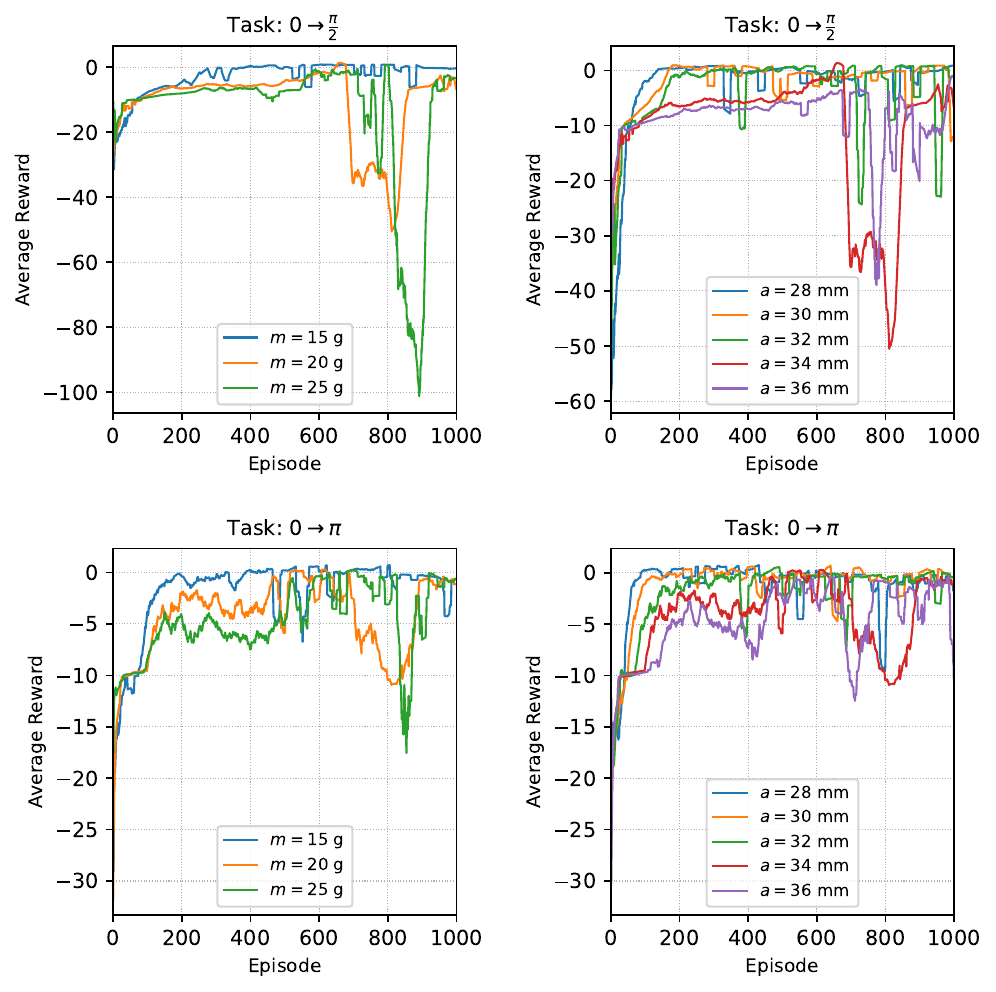}
\caption{Comparison of average reward during training for tasks $0 \to \frac{\pi}{2}$ and $0 \to \pi$ under different elliptical cylinder parameters: mass $m$ in grams and length of semi-major axis $a$ in millimeters. Smoothed with moving average of window=20 for all 1000 episodes.}
  \label{fig:training_reward_comparison}
\end{figure}

Fig. \ref{fig:training_reward_comparison} shows how the physical parameters of the simulation influence the rewards achieved during training. For the heavier and longer cylinders, the agents needed more training to achieve their goals (Tab. \ref{tab:best_training_results}), i.e. a positive average reward during an episode, which meant that they spent more time within the targeted area ( defined inside the reward function ) than outside it.

\begin{figure}[htbp]
\centering
\includegraphics[width=0.95\textwidth]{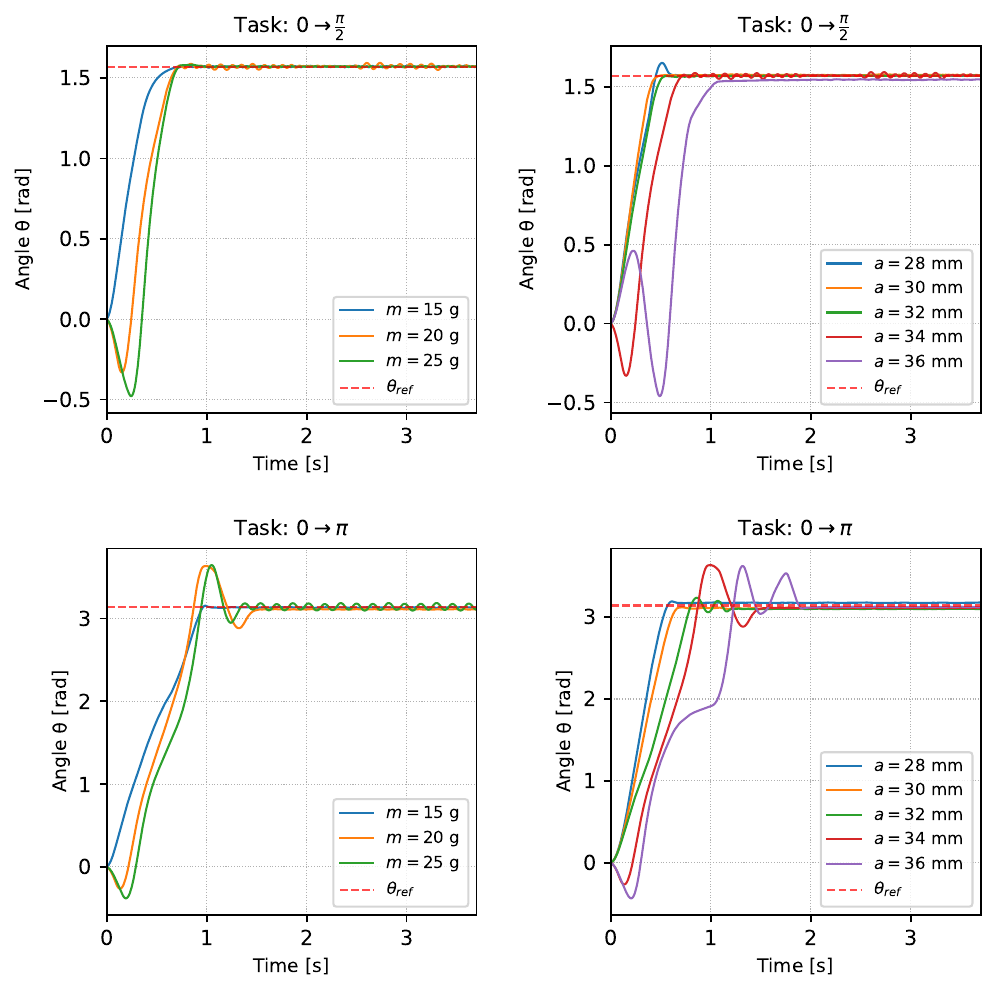}
\caption{Comparison of angular evolution for tasks $0 \to \frac{\pi}{2}$ and $0 \to \pi$ under different elliptical cylinder parameters: mass m in grams and
length of semi-major axis a in millimeters, controlled by the agent with the highest average reward recorded during training. 
  Each subplot shows the evolution of the angle $\theta$ as a function of simulation time. 
  The dashed red line denotes the reference angle $\theta_{\mathrm{ref}}$.}
  \label{fig:task_comparison1}
\end{figure}

In Fig. \ref{fig:task_comparison1}, different paths can be seen depending on the physical parameters of the cylinder. We take into account varying mass $m$ and varying semi-major axis $a$ keeping constant semi-minor axis $b$. For $m=15\ g$, for both tasks, the models did not have to accumulate potential energy in order to accomplish $\theta _{ref}$; the same goes for $a\le32 \ mm$. On the other hand, for tasks $0 \to \frac{\pi}{2}$ and $a=36 \ mm$, the agent performed a double swing. For task  $0 \to \pi$ with $a \ge 34 \ mm$ or $m\ge 20 \ g$, a struggle to maintain the $\theta_{ref}$ can be observed as it fluctuates further and back.

\begin{figure}[htbp]
\centering
\includegraphics[width=0.95\textwidth]{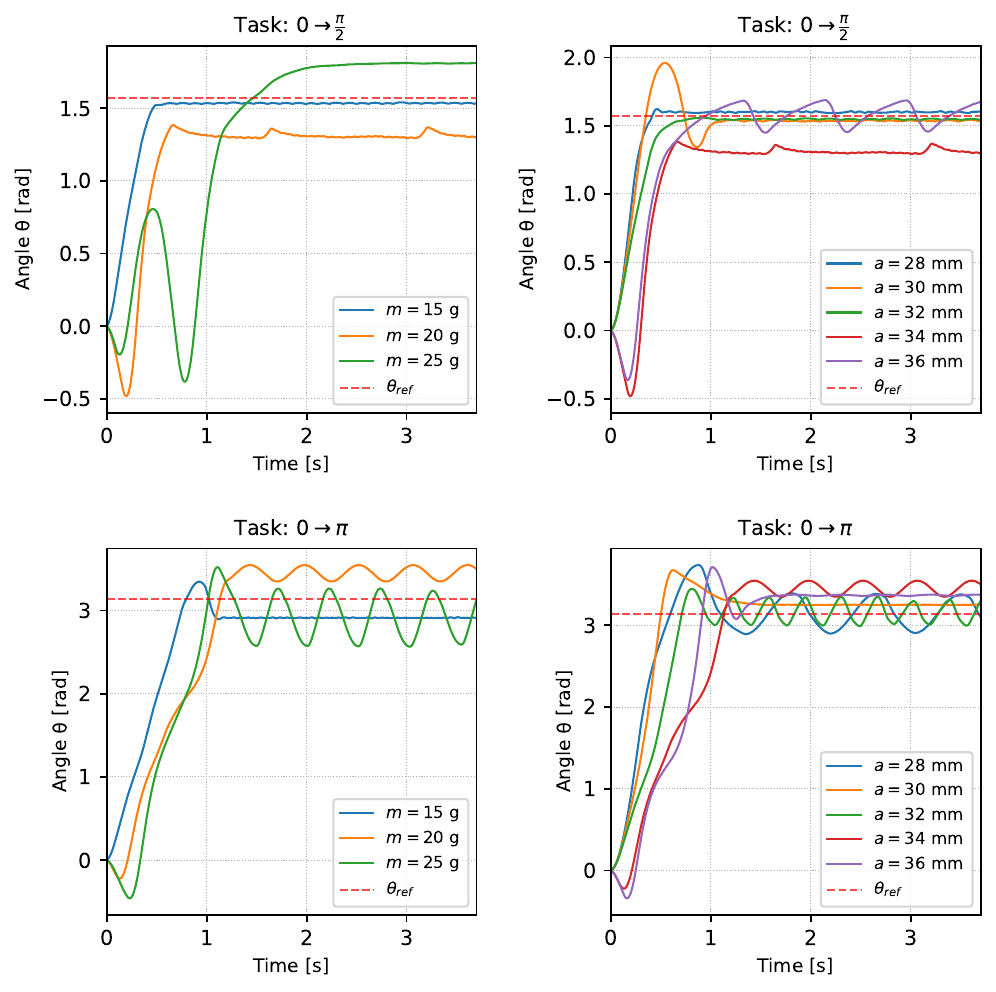}
\caption{ Similar to Fig. \ref{fig:task_comparison1} but controlled by agents trained for 1000 epochs and overatrained.}
  \label{fig:task_comparison_final}
\end{figure}

During training for experiments, early stopping was not applied in order to gather more data that are easier to compare. A backup of the model that achieved the highest positive average reward during 2048 steps (20.48 seconds) of training simulation was saved and used in experiments. Fig. \ref{fig:task_comparison_final} shows how the models behaved after exactly 1000 epochs of training, which, for most of the cylinder parameters, was more than enough; the models eventually became over-trained in comparison to Fig. \ref{fig:task_comparison1}, where the best models were chosen. 

\begin{table}[h]
\centering
\begin{tabular}{c|cc|cc}
\hline
& \multicolumn{2}{c}{$0 \to \frac{\pi}{2}$} 
& \multicolumn{2}{c}{$0 \to \pi$} \\
  & Epoch & Avg. Reward & Epoch & Avg. Reward \\
\hline

m=15 & 267 & 1.3359 & 156 & 0.8308 \\
m=20 & 653 & 1.6466 & 687 & 0.6614 \\
m=25 & 759 & 0.8233 & 723 & 0.5390 \\

a=28 & 123 & 1.0423 & 263 & 0.8270 \\
a=30 & 175 & 1.3669 & 549 & 0.8299 \\
a=32 & 185 & 1.0719 & 484 & 0.7561 \\
a=34 & 653 & 1.6466 & 687 & 0.6614 \\
a=36 & 989 & 0.7283 & 920 & 0.6349 \\

\hline
\end{tabular}
\caption{Epoch number for best average reward obtained during training for different scenarios and parameters.}
\label{tab:best_training_results}
\end{table}

\subsection{Swing Up + LQR controller as baseline comparison}
\label{sec:su_lqr}

To provide an interpretable reference, the Reinforcement Learning controller (denoted as \textbf{RL}) is compared with a classical two-stage baseline denoted as \textbf{SU} (swing-up + LQR). The control torque is bounded in all experiments, $\tau \in [-\tau_{\max},\tau_{\max}]$ with $\tau_{\max}=10^{-3}\,\mathrm{N\,m}$, and both RL and SU are evaluated using the same model parameters and actuation limits. The SU strategy follows a standard hybrid structure used for underactuated rotational systems: an energy-based swing-up stage drives the state toward the target region, and a local LQR stabilizer regulates the system near the desired equilibrium \cite{aastrom2000swinging,spong2002swing}.

As a baseline approach, the SU method combining energy shaping and a local LQR controller is considered. We define the total mechanical energy as
\begin{equation}
E(\theta,\omega) = \frac{1}{2} J(\theta)\,\omega^2 + V(\theta),
\label{eq:su_energy}
\end{equation}
where $J(\theta)$ and $V(\theta)$ is the potential energy corresponding to $\tau_p(\theta)$. 
The function $V(\theta)$ denotes the gravitational potential energy of the elliptical cylinder. It is defined such that its angular derivative corresponds to the gravity-induced torque, i.e.,
\begin{equation}
\tau_p(\theta) = \frac{dV(\theta)}{d\theta}.
\end{equation}
Therefore, $V(\theta)$ represents the configuration-dependent height of the center of mass of the cylinder and determines the equilibrium structure of the system. Stable equilibria correspond to local minima of $V(\theta)$, while unstable equilibria (e.g., the upright configuration) correspond to local maxima. This formulation allows the swing-up controller to regulate the total mechanical energy of the system toward the desired energy level
\begin{equation}
E^{\ast} = V(\theta_{\mathrm{ref}}).
\end{equation}

Treating the total energy as a control variable is consistent with the energy-shaping perspective and related passivity-based formulations \cite{ortega2002interconnection}.

During swing-up, the controller aims to match the current energy to the energy at the reference configuration. The target energy is set as:
\begin{equation}
E^\star = V(\theta_{\mathrm{ref}}),
\label{eq:su_estar}
\end{equation}
and the swing-up torque is defined as follows:
\begin{equation}
\tau_{\mathrm{su}}(\theta,\omega)=
\mathrm{sat}_{\tau_{\max}}
\Big(k_E\,(E^\star - E(\theta,\omega))\,\mathrm{sgn}(\omega)\Big),
\label{eq:su_law}
\end{equation}
where $k_E>0$ is a gain and $\mathrm{sat}_{\tau_{\max}}(\cdot)$ limits the torque to $[-\tau_{\max},\tau_{\max}]$. In all experiments, the swing-up gain was set to $k_E = 0.8$ and remained constant for all considered tasks and parameter variations (different masses $m$ and semi-major axes $a$). No individual retuning of $k_E$ was performed for specific cases, ensuring a consistent baseline comparison.
For very small $\omega$, the sign term is weakened to avoid fast switching, which is a common practical modification of energy-based swing-up laws \cite{spong2002swing}.

Close to the goal, we linearize the dynamics around $(\theta,\omega)=(\theta_{\mathrm{ref}},0)$ and define the local state $\mathbf{x}=[\tilde\theta\ \ \omega]^T$, where $\tilde\theta=\mathrm{wrap}(\theta-\theta_{\mathrm{ref}})$. The stabilizing torque is described as follows:
\begin{equation}
\tau_{\mathrm{lqr}}(\theta,\omega)=
\mathrm{sat}_{\tau_{\max}}
\Big(\tau_{\mathrm{eq}} - \mathbf{K}\,\mathbf{x}\Big),
\label{eq:lqr_law_simple}
\end{equation}
where $\mathbf{K}=[k_1\ k_2]$ is the LQR gain and $\tau_{\mathrm{eq}}$ compensates the static torque at the equilibrium. The gain matrix $\mathbf{K}$ is obtained by solving the algebraic Riccati equation for the linearized system \cite{anderson2007optimal,aastrom2021feedback}.

Finally, the commanded torque is formed as a smooth weighted combination of the swing-up and LQR terms:
\begin{equation}
\tau(\theta,\omega)=\alpha(\theta,\omega)\,\tau_{\mathrm{lqr}}(\theta,\omega)+\big(1-\alpha(\theta,\omega)\big)\,\tau_{\mathrm{su}}(\theta,\omega),
\label{eq:su_switch}
\end{equation}
where $\alpha(\theta,\omega)\in[0,1]$ increases toward 1 when $|\tilde\theta|$ and $|\omega|$ are small. In practice, $\alpha$ is implemented as a smooth ramp so that the controller transitions gradually from swing-up to LQR, avoiding a hard switch and reducing discontinuities in the control signal.

\begin{figure}[htbp]
\centering
\includegraphics[width=0.98\textwidth]{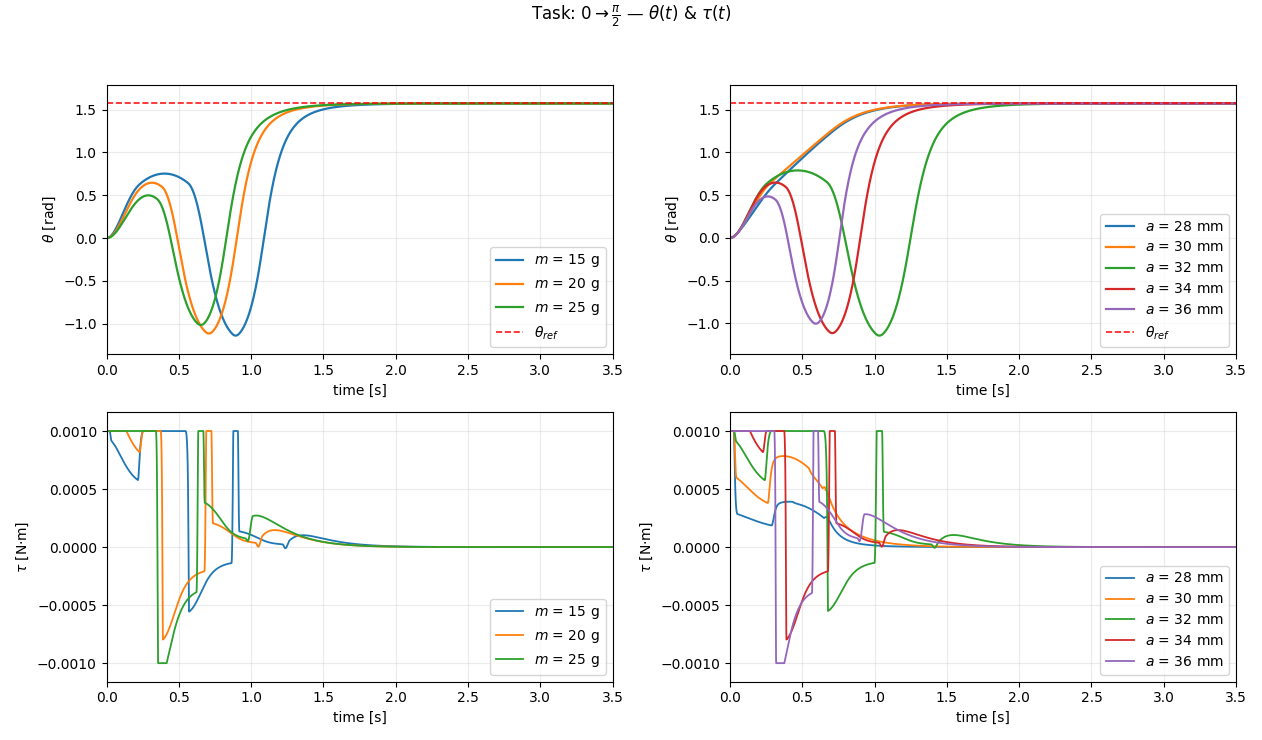}
\caption{SU (swing-up + LQR) baseline fortask $0 \rightarrow \frac{\pi}{2}$: angle response $\theta(t)$ (top row) and torque $\tau(t)$ (bottom row) for different masses (left column) and semi-axis values $a$ (right column). The dashed line indicates the reference angle $\theta_{\mathrm{ref}}=\frac{\pi}{2}$.}
\label{fig:ct_0_90}
\end{figure}

Fig.~\ref{fig:ct_0_90} shows the SU responses for task $0 \rightarrow \frac{\pi}{2}$: angle evolution $\theta(t)$ and control torque $\tau(t)$ for different masses and semi-axis values $a$. In all shown variants, the controller reaches the reference $\theta_{\mathrm{ref}}=\frac{\pi}{2}$ with a short swing-up transient, after which the torque decays toward zero as the stabilizer removes residual motion. Parameter changes mainly affect the transient: heavier or more elongated cylinders typically require more pronounced energy injection, which manifests as larger oscillations and brief torque saturation before settling.

\begin{figure}[htbp]
\centering
\includegraphics[width=0.98\textwidth]{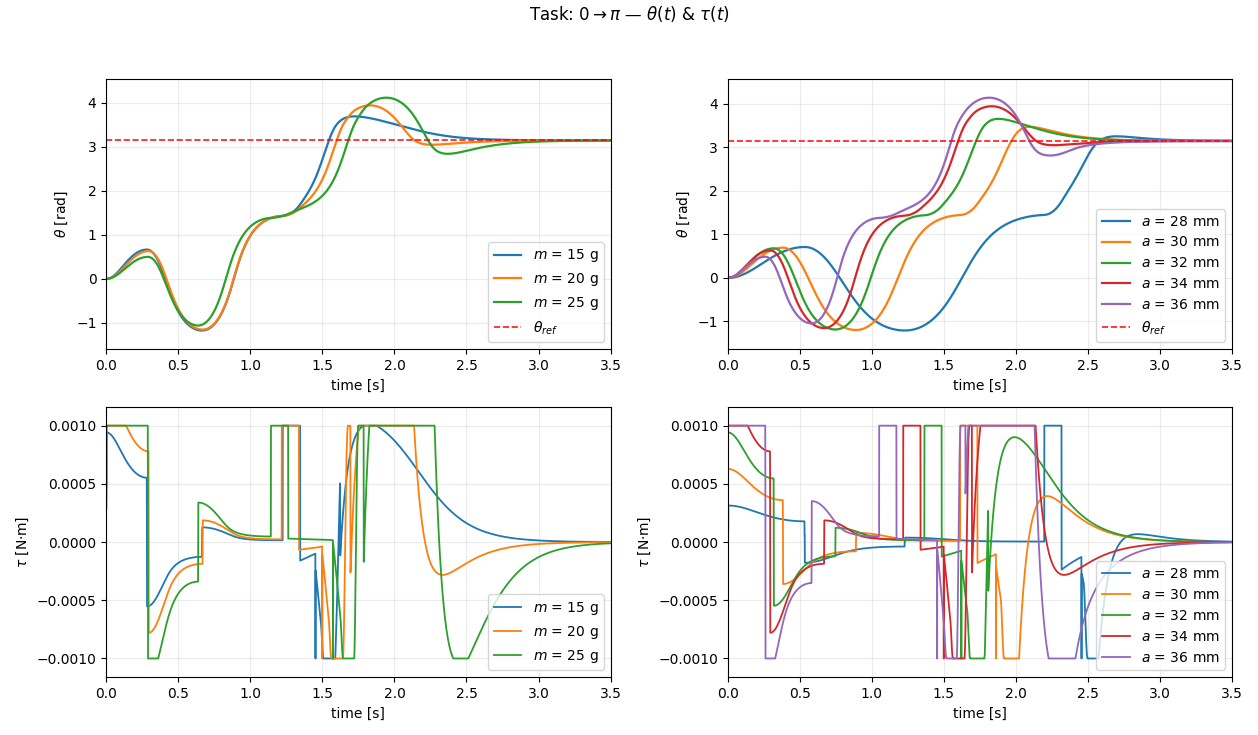}
\caption{SU (swing-up + LQR) baseline for task $0 \rightarrow \pi$: angle response $\theta(t)$ (top row) and torque $\tau(t)$ (bottom row) for different masses (left column) and semi-axis values $a$ (right column). The dashed line indicates the reference angle $\theta_{\mathrm{ref}}=\pi$.}
\label{fig:ct_0_180}
\end{figure}

The same analysis for task $0 \rightarrow \pi$ is presented in Fig.~\ref{fig:ct_0_180}. This scenario is more demanding, because the system must first accumulate enough energy to pass the upright configuration and then dissipate excess energy to converge to $\theta_{\mathrm{ref}}=\pi$. Consequently, the trajectories often exhibit a larger overshoot beyond the reference and a more complex torque pattern, including multiple saturated pulses during the transition and braking action near the goal. 

\begin{figure}[htbp]
\centering
\includegraphics[width=0.82\textwidth]{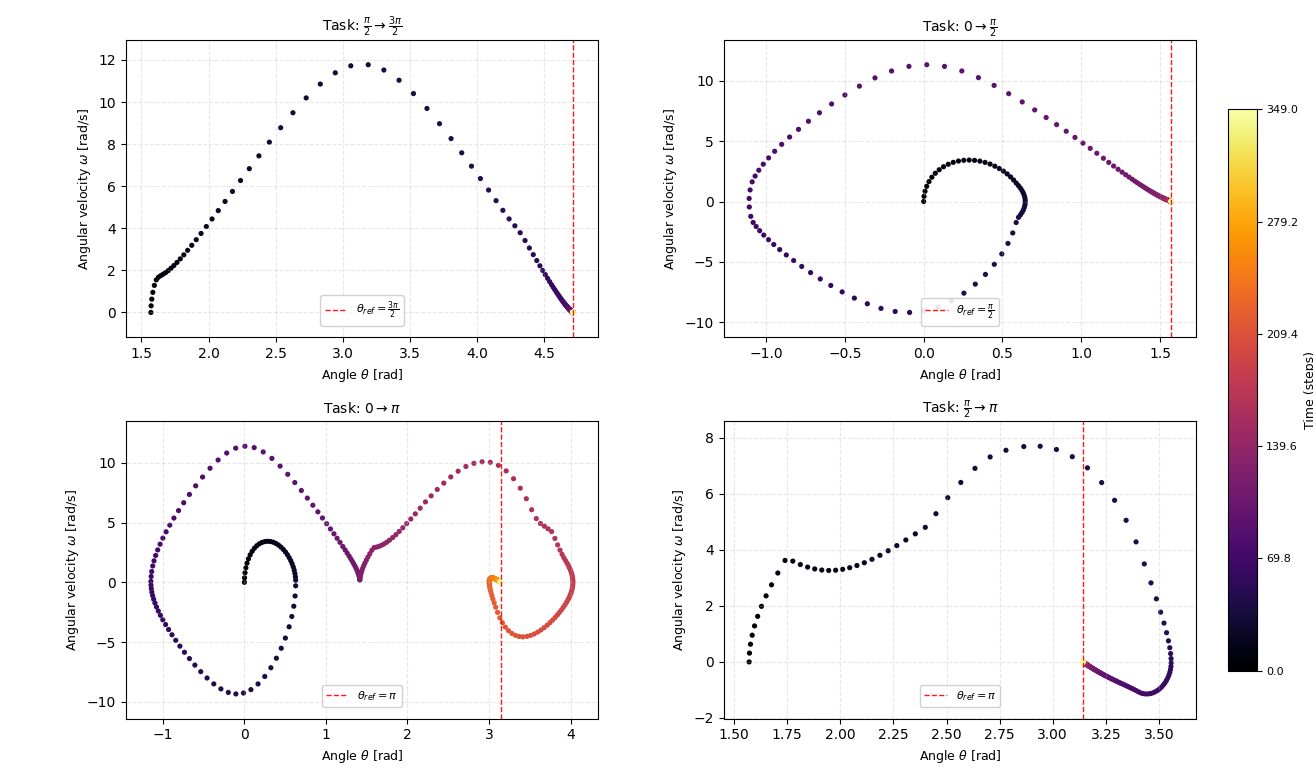}
\caption{Phase portraits $(\theta,\omega)$ for tasks $0 \rightarrow \pi$ and $0 \rightarrow \frac{\pi}{2}$ with step/time coloring. The dashed red line marks the reference angle $\theta_{\mathrm{ref}}$.}
\label{fig:ct_phase}
\end{figure}

To complement the time-domain plots, Fig.~\ref{fig:ct_phase} shows phase portraits $(\theta,\omega)$ with a step/time gradient. For $0 \rightarrow \frac{\pi}{2}$ the trajectory approaches the reference and then spirals inward as damping dominates. For $0 \rightarrow \pi$ an additional loop is visible, corresponding to crossing the unstable region near $\frac{\pi}{2}$ and subsequent removal of surplus kinetic energy before convergence.

The RL and SU are evaluated using ISE/ITSE-type criteria and step-response inspired metrics. Exactly, for tasks $0 \to \frac{\pi}{2}$ and $0 \to \pi$, the parameters ISE (Integral of the Square of the Error) and ITSE (Integral of Time multiplied by Square Error):
\begin{equation}
\label{eq:ISE}
ISE = \sum_{n=0}^{N} e^2(n) \Delta t
\end{equation}
\begin{equation}
\label{eq:ITSE}
ITSE= \sum_{n=0}^{N}t(n) \ e^2(n) \Delta t
\end{equation}
were collected in Tab. \ref{tab:ISE_ITSE}. In equations \ref{eq:ISE} and \ref{eq:ITSE}, $e(n)$ denotes the difference between the current angular position $\theta(n \Delta t)$ and the reference angular position $\theta_{\mathrm{ref}}(n\Delta t)$ for sample $n$, while $\Delta t$ denotes the sample time interval in a scenario with $N$ samples. The quantitative comparison reported in Tab.~\ref{tab:ISE_ITSE} and Tab.~\ref{tab:SU_RL_comp} clearly shows tha RL algorithm performs better than SU. Furthemore, for all cases the increase of mass decreases of the performance of RL algorithm. At the beginning, the increase of $a$ does not influence the performance, however for last two cases (a=$34$, $36$) where ratio of major/minor is high, the performance of RL strongly decreases. The SU algorithm is much more difficult to tune and in general performs much worse than RL.

\begin{table}[h]
\centering
\begin{tabular}{ccccccccc}
\hline
 & \multicolumn{4}{c}{final ISE $[rad^2s]$} & \multicolumn{4}{c}{final ITSE $[rad^2s^2]$} \\
 \hline
 method & $RL$ & $SU$ & $RL$ & $SU$ & $RL$ & $SU$ & $RL$ & $SU$\\
\cmidrule(r){2-5}
\cmidrule(r){6-9}
parameter & $0 \to \frac{\pi}{2}$ & $0 \to \frac{\pi}{2}$ & $0 \to \pi$ & $0 \to \pi$ & $0 \to \frac{\pi}{2}$ & $0 \to \frac{\pi}{2}$ & $0 \to \pi$ & $0 \to \pi$\\ \hline
m=15 & \textbf{0.39} & 3.22 & 3.05 & 12.78 & 0.04 & 2.42 & 0.70 & 7.89\\
m=20 & 1.00 & 3.09  & 4.54 & 12.87 & 0.17 & 1.86  & 1.14 & 8.10\\
m=25 & 1.39 & 2.85 & 5.49 & 12.55 & 0.30 & 1.54 & 1.56 & 7.98\\ \hline
a=28 & 0.40 & 0.70 & \textbf{2.16} & 22.43 & 0.04 & 0.15 & \textbf{0.30} & 24.29\\
a=30 & \textbf{0.39} & \textbf{0.65} & 2.42 & 16.60 & \textbf{0.038} & \textbf{0.14} & 0.39 & 13.28 \\
a=32 & 0.43 & 3.56  & 3.08 & 14.20 & 0.049 & 3.09 & 0.66 & 9.74\\
a=34 & 1.00 & 3.09 & 4.54 & 12.87 & 0.17 & 1.86 & 1.14 & 8.10\\
a=36 & 1.60 & 2.64 & 5.94 & \textbf{11.58} & 0.60 & 1.30 & 2.00 & \textbf{6.81}\\ \hline
\end{tabular}
\caption{ISE and ITSE parameters for tasks $0 \to \frac{\pi}{2}$ and $0 \to \pi$  depending on weight m, size a and the control method used (RL vs. SU).}
\label{tab:ISE_ITSE}
\end{table}

\begin{table}[htbp]
\centering
\begin{tabular}{cccc}
\hline
\multirow{2}{*}{parameter} & \multirow{2}{*}{task} & \multirow{2}{*}{swing-up} & reinforcement\\ 
 & & & learning\\ \hline
&  $\frac{\pi}{2} \to \frac{3\pi}{2}$  & 0.72 & 0.81 \\
settling time & $0 \to \frac{\pi}{2}$ & 1.45 & 0.68  \\
 {[s]} & $0 \to \pi$  & 2.43 & 1.46  \\
& $\frac{\pi}{2} \to \pi$ & 1.20 & 0.83  \\
\hline
& $\frac{\pi}{2} \to \frac{3\pi}{2}$ & 0.063 & 0.031  \\
error band& $0 \to \frac{\pi}{2}$ & 0.031 & 0.024 \\
{[rad]}& $0 \to \pi$ & 0.063 & 0.057 \\
& $\frac{\pi}{2} \to \pi$ & 0.031 & 0.013 \\
\hline
\end{tabular}
\caption{Comparison of swing-up and reinforcement learning efficiency for mass 20g and semi-major/minor axis a=34, b=26.}
\label{tab:SU_RL_comp}
\end{table}

\section{Conclusions}
\label{sec:conclusions}
To address the challenges related with the control of an elliptical cylinder, we create the controller based on Reinforcemenet Learning. It shows that it is possible to learn the object to move four various problems. Two of them which (vertical to vertical and vertical to horizontal) are simple to learn and two of them (horizontal to horizontal and horizontal to vertical) are very difficult due to non-trival solution. In general, this problem shows that RL easily learns how to release potential energy and laboriously learns how to accumulate it to perform control goal. The main challenge was to create policy that allows one to swing up the cylinder and stabilize it. We also show the influence of the perimeter on the learning process and the results. It shows that a high ratio of perimeter makes the problem much more difficult but still solvable. Further, to compare whether it is worth applying the reinforcement learning strategy, we compare our solution with the classical approach with swing up + LQR controller. In most cases, the performance of controller learned by RL is better in comparison to swing up + LQR which we treat as baseline solution. 

In future work, it is still worth improving the environment to get closer to reality, for instance taking into account viscous friction or details of untethered magnetic force. One of our drawbacks, which is required to solve in future, is the non-continuous torque, which is difficult to produce in a physical actuator. The next task is also to validate the controller in an experimental laboratory setup.

\section*{Acknowledgements}
This work was produced as a result of research project no. 2024/53/B/ST7/01540, funded by the National Science Centre Poland. 

\bibliographystyle{unsrt}  
\bibliography{references}

\end{document}